# Countering Inconsistent Labelling by Google's Vision API for Rotated Images

Aman Apte*[1], Aritra Bandyopadhyay*[1], K Akhilesh Shenoy*[1],
Jason Peter Andrews*[1], Aditya Rathod[1], Manish Agnihotri[1], Aditya Jajodia[1]

[1] Manipal Institute of Technology, Manipal-576104, India
* Authors with equal contribution.

**Abstract.** Google's Vision API analyses images and provides a variety of output predictions, one such type is context-based labelling. In this paper, it is shown that adversarial examples that cause incorrect label prediction and spoofing can be generated by rotating the images. Due to the black-boxed nature of the API, a modular context-based pre-processing pipeline is proposed consisting of a Res-Net50 model, that predicts the angle by which the image must be rotated to correct its orientation. The pipeline successfully performs the correction whilst maintaining the image's resolution and feeds it to the API which generates labels similar to the original correctly oriented image and using a Percentage Error metric, the performance of the corrected images as compared to its rotated counterparts is found to be significantly higher. These observations imply that the API can benefit from such a pre-processing pipeline to increase robustness to rotational perturbances.

## 1    Introduction

The recent surge in computational power has led to a widespread increase in research and development in the field of Machine Learning (ML) from both commercial and academic domains [1]. The success of ML algorithms deployed for Computer Vision tasks [2], has led to a surge in their use due to higher accuracy and more economical framework compared to their human counterparts, to the extent that there are speculations of ML algorithms taking over tasks that were traditionally thought of to be only capable of being done by a human. To catalyze the use of ML algorithms, companies like Microsoft, Amazon, Google etc. have created platforms with a layer of abstraction to simplify usability and incentivize deployment without explicitly having to train models. Google has its own framework for image analysis called the Cloud Vision API [3] built on the REST API architecture [4][5]. Though the Vision API has become more robust to input perturbations over the past few years, extensive testing has revealed that adversarial inputs can easily be used to spoof it into generating false or inconsistent labels [6] without the knowledge of the Black boxed algorithms that work in the back end of the API [7][8][9]. This highlights an imperative need to design models sturdier against perturbations both environmental and artificial in order to decrease vulnerability in real-world deployment scenarios such as autonomous driving [10], drones [11] and medical imaging [12] to name a few. Studies have shown that it is possible to deceive the ML algorithms through adversarial attacks by exploiting the type of data that is used as input [13] [14].

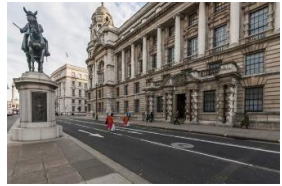

(a) Original Image (Label: Landmark)

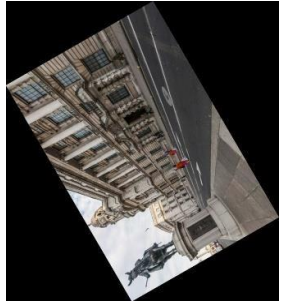

(b) Rotated 120 degrees (Label: Architecture)

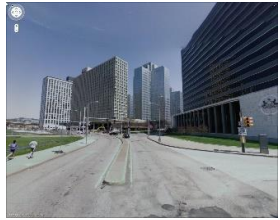

(c) Original Image (Label: Metropolitan Area)

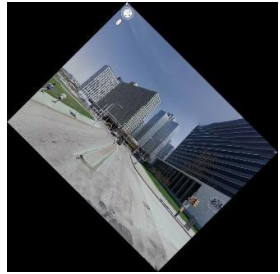

(d) Rotated 315 degrees (Label: Architecture)

**Fig. 1.** Illustration on mislabeling when Google's Cloud Vision API is fed rotated images as compared to the original images.

In this paper, the robustness of Google's Vision API to perturbations is evaluated, specifically, we investigate whether adversarial examples can be generated that would spoof the API but would be recognizable to a human.

The experiment revealed that after rotating the image by a sufficient angle, the API is misled into producing labels that are not aptly specific descriptors of the original image as seen in Figure 1 where only the highest confidence score labels are shown. As can be seen, the rotated images have inconsistent labels compared to their original counterparts. This variance with regards to rotation highlights the APIs vulnerability in physical world scenarios. For example, the object may be falsely classified when using satellite imagery. Moreover, the API can be attacked using such adversarial examples. Image filtering can be bypassed when fed rotated versions of inappropriate images [15].

Only having black-box access to the API, implementation of rotation invariance must be performed before feeding images to the API [16]. The results of our experiments revealed that when we perform context-based pre-processing on the rotated images, the API returns mostly the same labels with similar confidence scores as compared to the original image. These findings imply that the black boxed Vision API can benefit from image pre-processing without the need for retraining.

## 2    Related work

In view of the shortcomings of Machine Learning techniques in image analysis and lack of robustness to outdoor perturbations, there has been extensive research done on implementing rotational-invariant algorithms. The correction of image orientation has been performed in an adversarial environment using SVM [17]. It is also seen that using perception cues, image orientation correction is done using a Bayesian framework [18]. An image processing based approach also utilizes a histogram of radient orientations to determine orientation correction angle [19]. A Rotation-invariant orientation module [20] is proposed that can be used with existing popular CNN architectures. The incorporation of Rotational-invariant layers into existing Deep CNN architectures has been proposed for object detection [21]. The use of Radon Transform in a neural network [22] trained on unrotated images has been shown to correctly categorize rotated images. Local feature based rotationally-invariant Neoperceptrons [23] have been tested against standard Neoperceptrons and SIFT feature detection on rotated images. Local binary patterns used in conjunction with Logistic Regression [24] has been shown to have near human accuracy in regards to orientation detection. We intend to underline the susceptibility of misclassification by the black boxed API deployed in the real world against rotated inputs. We propose a CNN classification algorithm that performs orientation correction preprocessing of the input images for the Vision API.

Studies have shown that adversarial examples can be generated such that an ML algorithm performs misclassification whilst a human observer can understand the context [25]. This method can be further extrapolated to generate adversarial examples against black-boxed ML algorithms using only the output labels for given inputs [26]. To increase the robustness of the model through de-noising, a Deep Contractive Network [27] has been proposed. Taking into account the pitfalls of having adversarial inputs, an adversarial example detector [28] has been implemented and novel training techniques have been described to increase robustness in adversarial environments. Though a lot of work is being done in the field of robustness towards adversarial attacks, taking into account the success rate of such attacks [29], we must take these perturbations into account whilst designing deployable models.

## 3      Image rotation

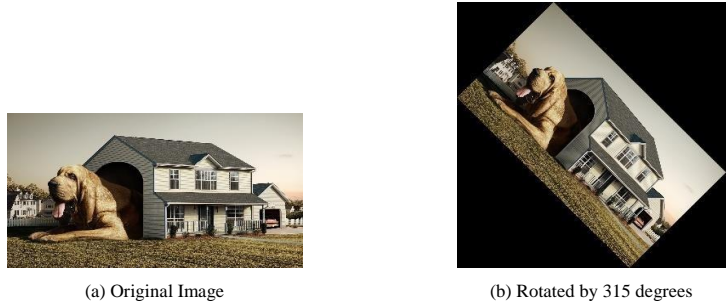

(a) Original Image                    (b) Rotated by 315 degrees

**Fig. 2.** Images rotated by specified angles.

In our implementation, we obtain a rotated image by passing the original image along with a rotation angle through a function that returns an image that contains the original image rotated counter-clockwise by the rotation angle about the center pixel with black padding. The rotated image within the padding has the same dimensions as the original image, thus retaining its original resolution.

We perform resizing on the rotated image in order to fit the input parameters of our deep learning model [30] which will then predict the angle by which the image has to be rotated to correct its orientation. After the image has been corrected using the predicted angle, we pass it through a function that recursively trims the black padding thus leaving only the correctly oriented original image with the original resolution. All the image processing has been done using the OpenCV library on Python.

## 4      Vision API test procedure

The testing procedure against Googles Cloud Vision API is explained in this section. For an input image, the API provides different types of output predictions such as context based labelling with a confidence score for each label, Optical Character Recognition, Face detection with sentiment prediction, provides metadata about the image and makes predictions about likeliness that the image contains content that is indecorous, including medical, spoof, adult or violent content. For our test procedure, we will be focusing on the APIs ability to provide context-based labels along with a confidence score for each label. We perform the test using street view images.

First, we input the original image into the API and store the output labels and corresponding confidence scores. These labels and scores will be used as the basis for comparing and analyzing the output labels and scores of the test images. We then rotate the original image from 0 to 360 degrees in increments of 3 degrees and input each iteration of the rotated original image into the API and store the labels and confidence scores. It is observed that for images rotated by small angles, the API outputs fewer high confidence score labels and for images rotated by large angles, the API fails to return most of the high confidence score labels. We then use this set of labels and confidence scores along with the original image's labels and scores to plot a graph of Percentage Error over rotation angles using the metric explained in the next section. The entire testing procedure was executed on a virtual machine hosted on the Google Cloud Platform [31] thus permitting us to test their Vision API extensively.

## 5 Similarity index metric

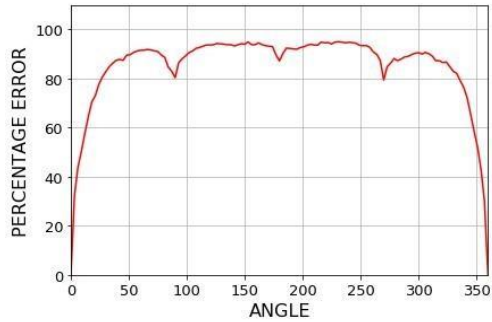

**Fig. 3.** The metric described below is used to plot the mean of the percentage error ratio for a set of 100 images each rotated from 0 to 360 degrees in steps of 3 degrees.

The Google Vision API analyses images by categorizing them with various labels that describe the image and by assigning confidence scores that represent the probability that the label assigned is correct. The metric system we have utilized works on a similar concept. After our pipeline corrects the rotated images to get the 'corrected image', the original and corrected images are compared on a 'Similarity Index' ranging from 0 to 100, where the higher the value, the more the similar the corrected image is to the original image.

The indexing values of the images are determined by assigning a particular weight to the API labels of that particular image that is proportional to the confidence score of the label. All the weights of the 'original image', hereafter taken as 'img1', are summed to equal the maximum index score, i.e., 100.

The weights are calculated as follows:

$$Weight(x) = (img1\ confidence(x) \div \sum_{\forall x \in img1} img1\ confidence(x)) \times 100 \qquad (1)$$

The indexing of the 'test image', hereafter taken as 'img2', is done in a similar manner wherein only the labels that are common to both: original and test images are considered for calculating the weights. The weights of the test image, hereafter called the 'Similarity Value', are calculated as follows:

If test image confidence(x) is less than the original image confidence(x) then,

$$Similarity\ Value(x) = Weight(x) \times (img2\ confidence(x) \div img1\ confidence(x)) \quad (2)$$

If test image confidence(x) is greater than or equal to the original image confidence(x) then,

$$Similarity\ Value(x) = Weight(x) \qquad (3)$$

Thus, the Similarity Index is calculated as follows:

$$Similarity\ Index = \sum_{\forall\ x\ \epsilon\ type1} Similarity\ Value(x) \qquad (4)$$

The percentage Error is obtained as follows,

$$percentage\ Error = 100 - Similarity\ Index \qquad (5)$$

The Similarity Index doesn't take into account labels from the test images' output that are contextually similar to the labels from the original image, making it a high punishment and zero rewarding metric, leading to a static baseline error in our rotation-correction model output.

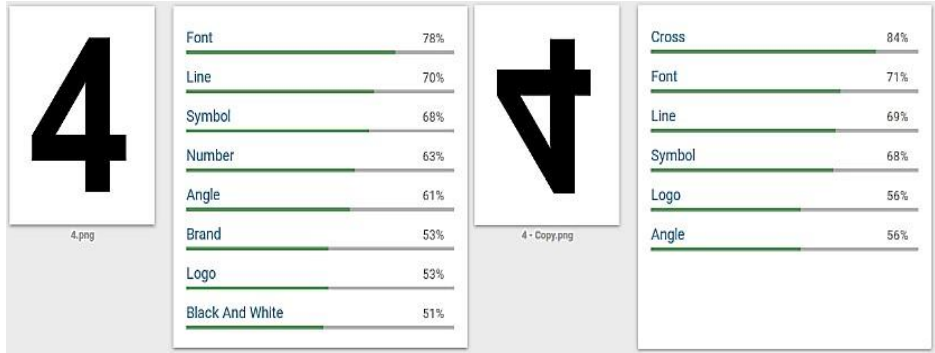

**Fig. 4.** Labels returned by the Vision API for which the Similarity Index is calculated below.

**1) Calculation of weights for img1 labels: (Using equation (1))**
  *Sum of all confidence* = 78+70+68+63+61+53+53+51 = 497
  *Weights:*
  Font = (78/497) *100 = 15.69        Line = (70/497) *100 = 14.08
  Symbol = (68/497) *100 = 13.69      Number = (63/497) *100=12.68
  Angle = (61/497) *100 = 12.28       Brand = (53/497) *100 = 10.66
  Logo = (53/497) *100 = 10.66        Black and White = (51/497) *100 = 10.26 **2)**
**2) Calculation of similarity values: (Using equations (2) and (3))**
  Labels common to img1 and img2 - Font, Line, Symbol, Logo, Angle.
  Font = 15.69 * (71/78) = 14.28      Line = 14.08 * (69/70) = 13.88
  Symbol = 13.69                      Logo = 10.66
  Angle = 12.28 * (56/61) = 11.27
**3) Calculation of Similarity Index: (Using equation (4))**
  Similarity Index =14.28+13.88+13.69+10.66+11.27 = 63.78
**4) Calculation of percentage Error: (Using equation (5))**
  percentage Error = 100 - 63.78 = 36.22

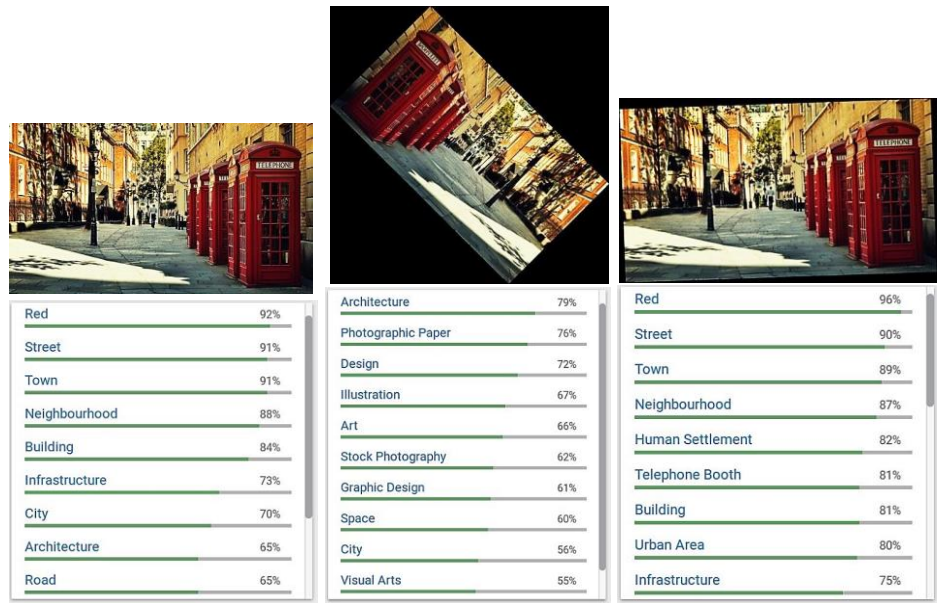

(a)APIs output labels for original Image (b) APIs output labels for rotated Image (c) APIs output labels for corrected image (45 Degrees)

**Fig. 5.** The labels returned by Cloud Vision API for original, rotated and corrected images. Some rotated image's labels are inconsistent compared to the labels of the original image, while the corrected image labels are mostly the same as the original image's labels.

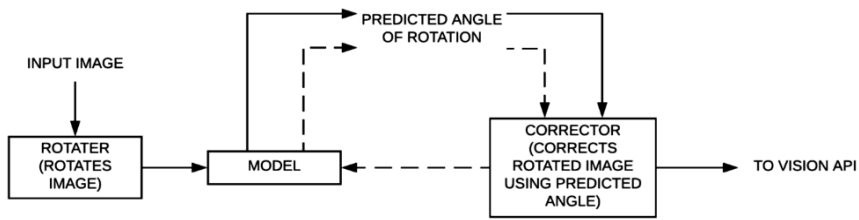

**Fig. 6.** The flowchart above illustrates the pipeline that we built and used to test the Google Vision API. The solid arrows represent the initial pass of the input image through the model and the dotted arrows represent the 2nd pass of the initially corrected image through the model.

## 6 Countermeasures

The error in labelling evident in Fig1 illustrates the importance of implementing rotation-invariant features in Computer Vision systems, leading to robustness to input perturbations, which makes it more deployable in fields like Satellite imagery and Military

Surveillance. Studies have shown that by performing data augmentation or regularization during training [32], we can improve the robustness of Machine Learning algorithms. It was also proposed that adversarial training (which creates and includes adversarial examples in the training data) could be another way to improve robustness. However, since the Vision API is a black-boxed algorithm, any approach which is based on robust optimization may not be viable.

Hence, a more modular approach would be to perform preprocessing of images before inputting them into the API. For this approach, we have used a ResNet50 architecture model which uses weights obtained from pre-training on the 'ImageNet' dataset. The weights are fine-tuned by further training using a set of 41,372 'street view' images for 50 epochs. To reduce computation and memory cost, we use a data-generator called "RotNetDataGenerator" which produces batches of rotated images and their respective rotation angles from a NumPy array of input images on-the fly [33].

The model has an output vector containing 360 classes (one for every angle) and predicts the angle by which an input image must be rotated to correct the orientation by outputting the class with the highest probability. Due to the orientation of features used to predict the angle, the model occasionally predicts angles such that the corrected image is a 180 degrees rotated version of the original image. To counter this problem, we have built the pipeline such that the corrected image gets fed back into the model and we use this output to get our final corrected image. The block representation of the preprocessing pipeline to the API is shown in Figure 6.

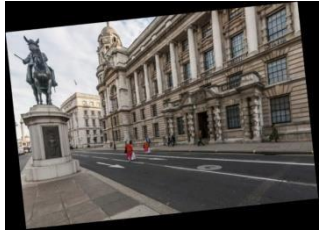 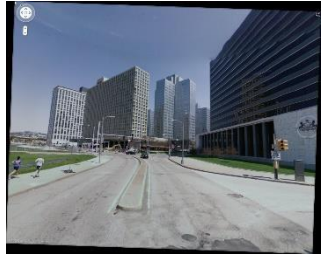

(a) Corrected Image (Angle 120 degrees)     (b) Corrected Image (Angle 315 degrees)

**Fig. 7.** The images rotated in the above sections, corrected after passing through the model, the API correctly predicts labels with similar confidence scores compared to the original image.

To test the accuracy of our model's predictions, for a test set of 100 images, we rotate every image from 0 to 360 degrees in increments of 3 degrees and input each rotated image into the pipeline and obtain the predicted angle. In Fig5, to assess the quality of angle correction, we plot the average of the difference between the rotated and predicted angle for every step angle.

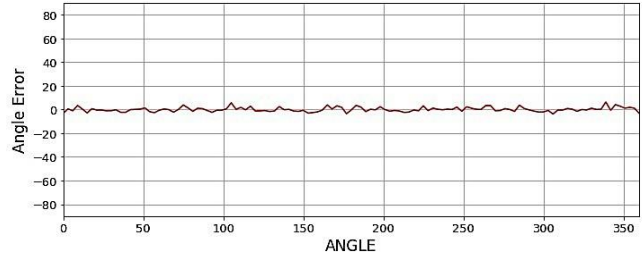

**Fig. 8.** This graph is to measure the accuracy of the angle correction predictions made by the model, it is the difference between the Rotated angle and Corrected angle for a set of 100 images from 0 to 360 degrees in steps of 3 degrees.

Using the same test set of rotated images described above, we plot the Percentage Error vs rotation angle for the images corrected using our pipeline, this graph (in blue) is overlayed on the graph plotted for rotated images (in red) to compare the reduction in Percentage Error for every step angle.

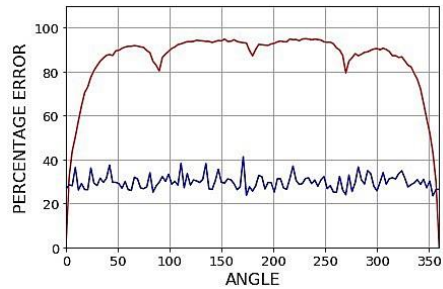

**Fig. 9.** A test set of 100 images is tested for rotations from 0 to 360 degrees in steps of 3 degrees where red is the plot of the rotated images and blue is the plot of corrected images.

Fig6. consists of the screenshots of the output labels of the API for original, rotated and corrected image. As we can see, the labels returned for the rotated image are inconsistent with the original image's labels and the labels of the corrected image are similar to the original labels.

Likewise, Fig4. shows corrected images of the images from Fig1. Unlike with the rotated images, the API correctly performs labelling with similar confidence scores on the corrected images, which suggests that application-specific black boxed Vision API tools could benefit from pre-trained rotation correction to invoke rotation invariance in the pipeline.

## 7    Conclusion

In this paper, we have shown that even without explicit knowledge of the black-boxed algorithm, adversarial examples can be generated to spoof the Vision API by rotating images, which results in inconsistent labelling by the API. This highlights the API's lack of invariance towards rotated inputs. We also proposed a modular orientation correction pre-processing pipeline using a ResNet50 model and evaluated its accuracy and the similarity between the original and corrected image's outputs from the API. We find that the API generates mostly similar labels for the corrected image as compared to the original image. Thus, also taking into account the extensive research being done on rotational-invariance in Computer Vision, we can observe that the API could benefit from such a preprocessing pipeline to improve robustness and increase the scope of deploy ability in real-world scenarios.